\documentclass{article} 
\usepackage{style,times}

\usepackage{amsmath,amsfonts,bm}

\def\eqref#1{equation~\ref{#1}}

\def\1{\bm{1}}

\DeclareMathAlphabet{\mathsfit}{\encodingdefault}{\sfdefault}{m}{sl}
\SetMathAlphabet{\mathsfit}{bold}{\encodingdefault}{\sfdefault}{bx}{n}

\usepackage{hyperref}
\usepackage{url}
\usepackage{graphicx}
\usepackage{booktabs}
\usepackage{amsmath}
\usepackage{amssymb}
\usepackage{xcolor}
\usepackage{tikz}
\usepackage{longtable}
\usepackage{array}
\usepackage{xcolor}
\usepackage{tabularx}

\usetikzlibrary{arrows.meta,positioning,fit,backgrounds}

\newcommand{\draftmode}{1} 
\newcommand{\draftnote}[1]{\ifnum\draftmode=1\textcolor{red}{#1}\par\fi}

\title{EHRAdapt: Adapting Pretrained Language Models to Electronic Health Records with Semantic Priors for Rare Clinical Events}

\author{
Andre R.\ Goncalves\textsuperscript{1},
Vincent Liu\textsuperscript{2},
Priyadip Ray\textsuperscript{1}\\[0.35em]
\textsuperscript{1}Lawrence Livermore National Laboratory\\
\textsuperscript{2}Kaiser Permanente Division of Research
}

\preprinttrue

\begin{document}

\maketitle

\begin{abstract}
Electronic health records (EHRs) encode clinical histories as (time, modality, code) tuples, whereas pretrained language models expect text tokens. Serializing each tuple as text inflates sequence length and repeatedly encodes the same structure. We introduce EHRAdapt, an adapter that maps tuples directly into a frozen language model's embedding space. Modality receives a learned embedding, time gaps enter through learned attention biases, and event codes receive dedicated vectors.
Learning these event vectors is the central challenge: clinical vocabularies are long-tailed, leaving rare events with too few observations to estimate reliable representations independently. 
EHRAdapt therefore represents each event vector as the sum of a \textit{semantic prior} and an \textit{evidence residual}. The prior is a frozen embedding of the event's clinical description, from a biomedical language model trained on clinical ontologies, so it carries clinical knowledge. A shared learned projection maps this embedding into the model's input space, allowing the prior to supply clinical meaning even when observations are scarce.
The residual is a learned low-rank, event-specific correction that refines this representation as evidence accumulates.
We perform continued pretraining on records from about 4 million patients with three pretrained LLMs as backbones: OLMo2 1B, Llama3.2 1B, and OLMo2 7B. As the LLMs remain frozen, only the adapter is trained, which amounts to 0.1--0.6\% of all parameters.
The full adapter outperforms all ablations in held-out next-event prediction on every backbone. Removing the semantic pathway hurts rare events most, over ten times more than the most frequent ones, whereas removing the residual hurts overall prediction but improves it for the rarest events.
On reportable infectious-disease and syndromic downstream classification tasks, EHRAdapt outperforms text-based LLM and count-based baselines, as well as the same LLMs adapted to the same EHR records through next-event prediction, and both pathways improve rare-disease discrimination. The semantic prior and the evidence residual therefore play complementary roles, and these roles become visible only when results are broken down by event frequency rather than averaged.
\end{abstract}

\section{Introduction}

Pretrained large language models (LLMs) offer an attractive starting point for modeling structured electronic health records (EHRs). Foundation models built specifically for EHRs are trained from scratch on large patient corpora \citep{wornow2023ehrshot, waxler2025generativemedicaleventmodels}, and this costly training must be repeated to benefit from newer architectures. Adapting an existing LLM requires less data and computing infrastructure, and newer LLMs can be adopted as they are released. Recent work has therefore adapted pretrained LLMs to clinical prediction from structured EHRs by typically converting each clinical event into text \citep{ben2024cpllm,hegselmann2025ehrencoders,goncalves2025biosurveillance,wang2026integrating}. However, standard LLM tokenizers break a clinical event into multiple tokens (about 27 per event in our text serialization; Appendix~\ref{app:nep-ehradapt-training}), increasing the length of patient histories, while information about the event type (e.g., diagnosis, laboratory test, or medication) and its timing is encoded only implicitly.

Clinical events also follow a highly skewed distribution \citep{bates2010distribution}. A small set of routine events occurs very often, while most events are rare. In our training data, the median event type appears about 20 times, whereas the most common appears about 87.6 million times (Figure~\ref{fig:kp-long-tail}). Rare events may still carry important diagnostic information, but a model that learns each event only from EHR data sees too few examples of them. The skew also complicates evaluation: the average prediction loss is dominated by common events, so it can hide large effects on rare ones.

In this paper, we introduce EHRAdapt, an adapter that maps structured EHR events into the input space of a frozen pretrained LLM. Each event is represented by a single token, with event type encoded through learned embeddings and temporal information through attention biases. As the LLM backbone remains frozen, EHRAdapt is model-agnostic, and only the adapter is trained, with less than 1\% as many parameters as the backbone. To better represent rare events, EHRAdapt combines a semantic prior with an event-specific residual. The semantic prior is a BioLORD embedding of the event's clinical description \citep{remy2024biolord}, projected into the LLM input space by a learned map shared across events. Because BioLORD is trained on clinical ontologies such as the Unified Medical Language System (UMLS), this prior is informative even for rare events. 

We expect the contributions of the semantic prior and event-specific residual to depend on event frequency. Rare events should remain closer to their semantic priors due to limited observations, while common events can adapt more strongly through their learned residuals. This resembles Bayesian learning, where prior information has greater influence under data scarcity, although EHRAdapt neither performs Bayesian inference nor explicitly enforces this behavior. 

We pretrain EHRAdapt on records from 4 million Kaiser Permanente Northern California  patients collected between 2010 and 2018 \citep{goncalves2025biosurveillance}. We use three frozen backbones, OLMo 2 1B and 7B \citep{olmo2025olmo2} and Llama 3.2 1B \citep{grattafiori2024llama3}, and on each one we compare the full model with ablations of the semantic prior, event-specific residual, and learned semantic mapping. Our evaluation asks three questions. First, how does each part affect prediction of the next clinical event, and for which events? To answer this, we report results separately for ten frequency groups, from the rarest events to the most common. Second, do the learned representations improve downstream prediction for patients seen between 2019 and 2022? Third, does the balance between the two parts shift with event frequency, as we expect?

The downstream tasks, reportable infectious disease and syndrome classification, are motivated by biosurveillance, where earlier diagnosis can speed outbreak response \citep{henning2004syndromic, lazarus2009esp, goncalves2025biosurveillance}. Many diseases of surveillance interest are rare (Appendix~\ref{app:downstream-counts}), making these tasks a natural test of whether the semantic prior helps most where data are scarce. Rare diseases and rare events are related but not identical, so we evaluate both. 

Overall, EHRAdapt achieves the lowest test loss across all three backbones, with the semantic prior and event-specific residual playing complementary roles (Section~\ref{sec:results}). Removing the semantic prior degrades prediction across all frequencies, with the largest effect on rare events. In contrast, removing the residual improves prediction for the rarest events but degrades overall performance, suggesting that residuals primarily benefit common events. Consistently, residuals contribute less to representations of rare events than common events (Section~\ref{sec:representation}). On downstream tasks, EHRAdapt outperforms text-based LLM inputs, BioLORD embeddings alone, a tuned random forest, and the same backbones adapted to the same EHR corpus through next-event prediction, with both components improving rare-disease prediction (Section~\ref{sec:downstream}). 

\section{Related work}
Representation learning for structured EHRs has evolved from BERT-style models such as BEHRT, Med-BERT, and CEHR-BERT \citep{li2020behrt,rasmy2021medbert,pang2021cehrbert} to large-scale foundation models including CLMBR, TransformEHR, MOTOR, and Curiosity \citep{steinberg2021clmbr,yang2023transformehr,steinberg2024motor,waxler2025generativemedicaleventmodels}. These approaches learn reusable patient representations through pretraining on large longitudinal EHR corpora. A complementary direction adapts existing pretrained LLMs to structured EHR sequences through fine-tuning or prompting, reducing the need for EHR-specific pretraining and facilitating the adoption of newer models \citep{ben2024cpllm,zhu2024prompting,goncalves2025biosurveillance}. However, the long-tailed distribution of clinical events provides limited observations for learning robust representations of rare events. Prior work such as GRAM  \citep{choi2017gram} addresses this challenge by leveraging hierarchical relationships among medical concepts, constructing representations by attending over concept ancestors in a medical ontology. In contrast, EHRAdapt can incorporate semantic priors from arbitrary external sources, including biomedical language models and knowledge graphs, and adaptively balances these priors with representations learned from EHR data according to the available evidence for each clinical event.

Recent approaches use external semantic knowledge to represent clinical events. Biomedical ontologies and representation models such as BioLORD encode semantic relationships among clinical concepts independently of their frequency in a particular EHR corpus \citep{bodenreider2004umls,remy2022biolord,remy2024biolord}. Most closely related to our work, PORTER \citep{guo2026porter} represents clinical events using natural-language descriptions encoded by a frozen BioLORD encoder, enabling vocabulary-independent representations and improved transfer across coding systems and institutions. Whereas PORTER uses fixed semantic representations primarily for portability, EHRAdapt addresses the complementary problem of data scarcity for rare events through frequency-adaptive integration of semantic priors: rare events rely more on pretrained semantic knowledge, while representations of frequently observed events are increasingly shaped by evidence from the EHR corpus.

A separate line of work adapts LLMs to clinical notes and biomedical literature text. Because these documents are already written in natural language, they can be fed directly to a pretrained language model, and models such as Bio-ClinicalBERT, GatorTron, and NYUTron \citep{alsentzer2019clinicalbert, yang2022gatortron, jiang2023nyutron} are pretrained or fine-tuned on large collections of notes for tasks ranging from concept extraction to readmission and mortality prediction. Structured EHRs pose a different problem: clinical events are recorded as codes with timestamps rather than text, so they must first be mapped into a form an LLM can process. EHRAdapt focuses on this structured setting and is complementary to models of clinical notes.

\section{Method}
\label{sec:method}

\begin{figure}[t]
\centering
\includegraphics[width=\linewidth]{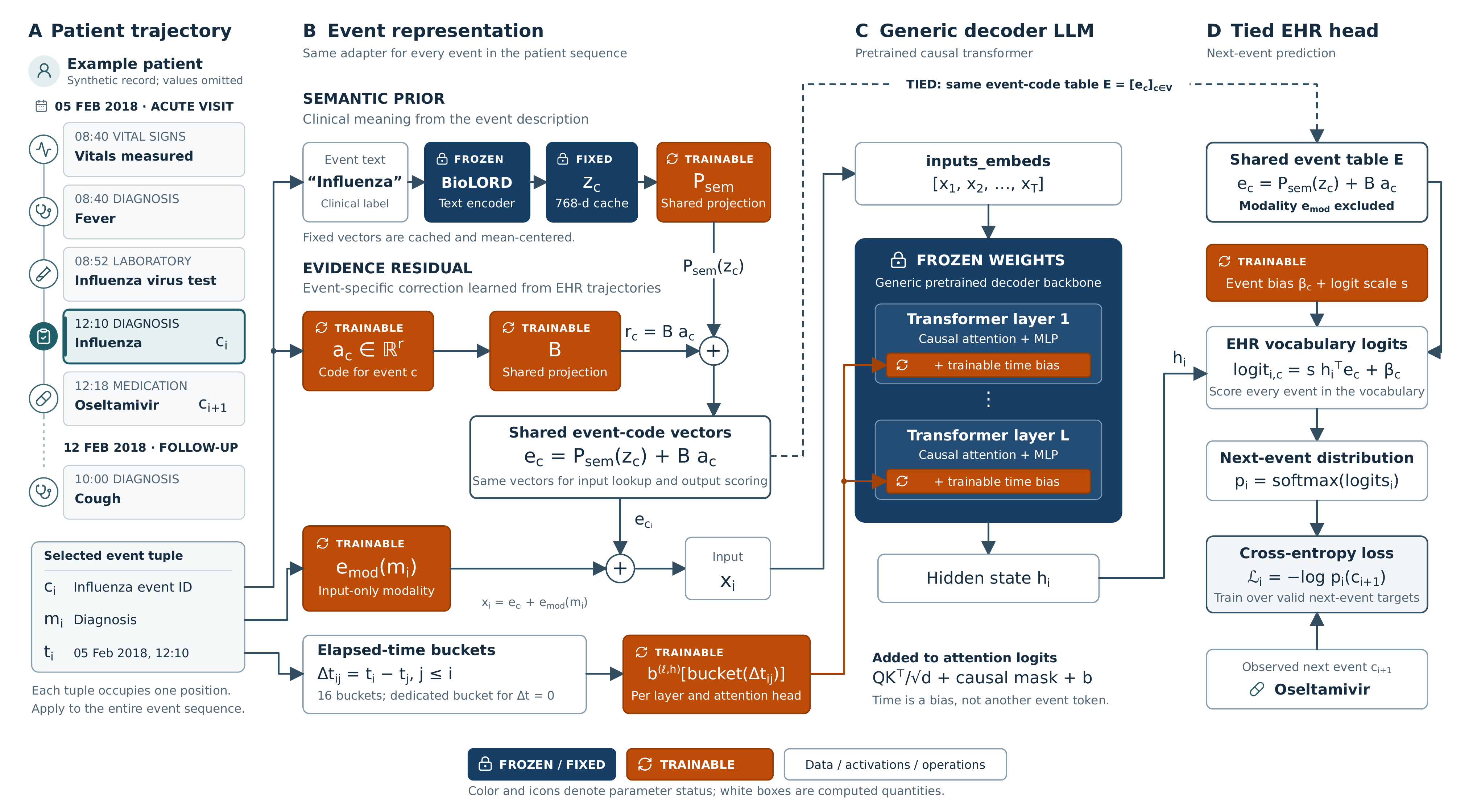}
\caption{\textbf{EHRAdapt CPT architecture.}
\textbf{(A)} Synthetic patient trajectory of timestamped, typed clinical events. \textbf{(B)} Frozen BioLORD encodes an event's clinical description as the semantic vector $\mathbf{z}_c$. A shared learned projection gives the semantic prior $P_{\mathrm{sem}}(\mathbf{z}_c)$, which is added to the low-rank evidence residual $\mathbf{B}\mathbf{a}_c$ to form the event embedding $\mathbf{e}_c$. 
\textbf{(C)} A frozen decoder LLM receives these vectors through \texttt{inputs\_embeds}. Pairwise elapsed-time buckets index learned attention biases per layer and head.
\textbf{(D)} The tied output head reuses $\mathbf{e}_c$ without modality, with a learned event bias $\beta_c$ and logit scale $s$; CPT minimizes next-event cross-entropy. Dark blue (lock): frozen; orange (update): trainable; white: data or computed.}
\label{fig:arch}
\end{figure}

EHRAdapt operates in two stages. First, its parameters are learned through continued pretraining (CPT) on next-event prediction over longitudinal EHR data. Second, the adapted model is reused across downstream tasks. In the simplest setting, it stays frozen and only a task-specific head is trained; alternatively, while not explored in this paper, parameter-efficient methods such as LoRA \citep{hu2022lora} can further adapt it to each task. Figure~\ref{fig:arch} summarizes the CPT architecture, from structured patient trajectories to next-event prediction. The following subsections describe each component.

\subsection{Problem setup and sequence construction}

A patient history is a sequence of clinical events $(c_1, m_1, t_1), \ldots,
(c_L, m_L, t_L)$, where $c_i$ indexes a live event in a vocabulary $\mathcal{V}$ of clinical event types plus special tokens, $m_i$ is the event modality, and $t_i$ is a timestamp. 
In the continued pretraining phase, the model is trained to predict $c_{i+1}$ given the prefix, so the objective is next-event cross-entropy in nats.


\subsection{Event representation}
\label{sec:event_representation}

For clinical event $c$, we distinguish the shared event-code vector
$\mathbf{e}_c$ from the modality-augmented input $\mathbf{x}_i$ at sequence
position $i$:
\begin{equation}
\begin{aligned}
\mathbf{e}_c &= P_{\mathrm{sem}}(\mathbf{z}_c) + \mathbf{r}_c,
\qquad \mathbf{r}_c = \mathbf{B}\,\mathbf{a}_c,
\quad \mathbf{a}_c \in \mathbb{R}^{r}, \\
\mathbf{x}_i &= \mathbf{e}_{c_i} + \mathbf{e}_{\mathrm{mod}}(m_i),
\end{aligned}
\label{eq:repr}
\end{equation}
where $\mathbf{z}_c \in \mathbb{R}^{d_{\text{sem}}}$ is the frozen BioLORD
embedding \citep{remy2024biolord} of the event name; $P_{\mathrm{sem}} \in \mathbb{R}^{H\times d_{\text{sem}}}$ is a learned map to the backbone hidden dimension $H$; $\mathbf{a}_c$
is a learned event-specific code of dimension $r$; $\mathbf{B} \in
\mathbb{R}^{H \times r}$ is a shared projection; and $\mathbf{e}_{\mathrm{mod}}(m_i)$ is a learned modality embedding, from a lookup table with one row per clinical modality and an additional row for special tokens.

\paragraph{Residual rank} The rank $r$ controls the capacity of the event-specific correction. Stacking the codes $\mathbf{a}_c$ into $\mathbf{A} \in \mathbb{R}^{|\mathcal{V}| \times r}$, where $|\mathcal{V}|$ is the vocabulary size, gives the residual table
$\mathbf{R} = \mathbf{A}\mathbf{B}^{\top}$ with
$\operatorname{rank}(\mathbf{R}) \leq r$. The factorization contains $(V+H)r$
parameters, compared with $VH$ for an unconstrained residual table. For a
backbone with $H = 2{,}048$, rank 64 uses approximately 5.15 million residual
parameters versus 161 million for an unconstrained table, about 3.2\% as many,
excluding the semantic projection and other adapter components. Higher ranks relax the constraint, and any residual table can be expressed once
$r \geq H$. However, greater representational capacity does not guarantee
better held-out performance, particularly for rare events.


\paragraph{Fixed semantic geometry and the prior interpretation.}
As an alternative to learning $P_{\mathrm{sem}}$, the projection can be
fixed at $P_{\mathrm{sem}} = \alpha Q$, where
$Q \in \mathbb{R}^{H \times d_{\text{sem}}}$ satisfies $Q^\top Q = I$ and
$\alpha > 0$. Consequently, $\cos(\alpha Q\mathbf{z}_h, \alpha Q\mathbf{z}_l) = \cos(\mathbf{z}_h, \mathbf{z}_l)$, so all angles in the BioLORD space are preserved exactly, and cosine-neighbor rankings are preserved. Combined with the
residual, this fixed projection corresponds most closely to a fixed-prior
interpretation of Equation~\ref{eq:repr}: the external semantic structure is
held constant while the residual absorbs evidence from EHR sequences.
Learning $P_{\mathrm{sem}}$, as in the default model, relaxes this
constraint and allows the semantic component to adapt to the backbone.
We evaluate the fixed projection, with and without the residual, as controls in Section~\ref{sec:results}.

The decomposition in Equation~\ref{eq:repr} is central to our analysis,
because its two pathways receive evidence differently. The semantic pathway
is shared across the whole vocabulary and is therefore optimized
predominantly by frequent events, but it reaches rare events through the
BioLORD geometry: a rare event's $\mathbf{z}_c$ lies near the $\mathbf{z}$ of
semantically related frequent events, so updates to $P_{\mathrm{sem}}$
driven by frequent events also move the rare event's representation. The
residual pathway instead assigns each event its own learned code
$\mathbf{a}_c$, while sharing the projection $\mathbf{B}$. 


\subsection{Output head, frequency bias, and logit scale}
\label{sec:head}
The output head is tied to the event-code representation table
$\mathbf{E} = [\mathbf{e}_c]_{c \in \mathcal{V}}$
\citep{press2017tied, inan2017tying}. For hidden state $\mathbf{h}_i$,
\begin{equation}
\mathrm{logit}_{i,c} \;=\; s\,\bigl(\mathbf{h}_i \cdot \mathbf{e}_c\bigr) + \beta_c,
\qquad s = \exp(\texttt{logit\_scale}),
\end{equation}
where $\beta_c$ and \texttt{logit\_scale} are learned and the modality
embedding is deliberately excluded from the output representation. With a
linear $P_{\mathrm{sem}}$ and a projected residual, the unscaled scores before
adding $\beta_c$ factorize as $(\mathbf{h}_i P_{\text{sem}})\mathbf{Z}^{\!\top} +
(\mathbf{h}_i\mathbf{B})\mathbf{A}^{\!\top}$, which avoids materializing the
$V \times H$ matrix and avoids constructing a dense event table for every forward pass.

Because the input and output embeddings are tied, each residual $\mathbf{a}_c$ receives gradients from two sources. On the input side, it is updated only when event $c$ appears in a patient history. On the output side, the full softmax scores every event at every position, so $\mathbf{a}_c$ is updated at every step, even when event $c$ is absent. Event frequency therefore determines how often $\mathbf{a}_c$ is updated with event $c$ present, as an input or as the prediction target, but not how often it is updated overall.


\subsection{Temporal attention bias}

Elapsed time carries clinical information that event order does not, since
consecutive events may be minutes or years apart. We therefore add a learned bias
to the attention scores. For query position $i$ and key position
$j \leq i$, with $\Delta t_{ij} = t_i - t_j$, the score at layer $\ell$ and
head $h$ is
\begin{equation}
\mathbf{q}_i^{\top}\mathbf{k}_j / \sqrt{d_{\text{head}}}
+ b^{(\ell,h)}_{\kappa(\Delta t_{ij})},
\label{eq:temporal}
\end{equation}
where $\kappa$ assigns elapsed time to one of 16 buckets, with bucket 0
for simultaneous events. This follows the learned, bucketed
relative-position biases of T5 \citep{raffel2020t5}, 
but uses elapsed time instead of token distance. Biases are initialized to zero, leaving the backbone's attention unchanged at the start of training.

\section{Experimental setup}
\label{sec:setup}

We learn the EHRAdapt parameters by continued pretraining (CPT), then
evaluate next-event prediction and transfer to two biosurveillance-related
diagnostic prediction tasks. The full adapter and four ablations are trained on three
backbones, described below, allowing us to distinguish component effects that recur across
models from differences associated with backbone family or size.

\subsection{Continued-pretraining data}
\label{sec:cpt-data}

We use Kaiser Permanente Northern California (KPNC) EHRs from approximately 4 million patients collected between 2010 and 2018. In addition to demographics, the data include
longitudinal records of laboratory tests, procedures, medications, encounter
types, chief complaints, diagnosis codes, and vital signs, without associated
values or outcomes. A deterministic, seeded patient-level split assigns 98\%,
1\%, and 1\% of patients to training, validation, and test sets.

Patient trajectories are partitioned into chunks of 16 to 512 events, with
boundaries aligned to timestamp changes where possible so that simultaneous
events are not split. A trailing chunk shorter than 16 events is merged into
its predecessor, so all events are retained. Each training epoch contains approximately 721 million prediction targets. The vocabulary, frequency strata, and output-bias initialization are derived from the training split only. See Appendix~\ref{app:data} for additional information.

\subsection{Backbones and configurations}
\label{sec:comparison}


We use frozen OLMo-2 1B and 7B backbones \citep{olmo2025olmo2}, which are
pretrained separately and provide a within-family size comparison, and
Llama 3.2 1B \citep{grattafiori2024llama3}, which provides a cross-family
comparison at the same hidden size. We do not include Llama 3.1 8B because
Llama 3.2 1B was derived from it by pruning and knowledge distillation, so
the pair would confound model size with model derivation. Additional backbone
information is in Appendix~\ref{app:sec_backbone_models}.

On each backbone, we compare the full adapter with four configurations that
remove or fix components of Equation~\ref{eq:repr}. The \emph{Full model}
learns $P_{\mathrm{sem}}$ and the rank-64 residual. \emph{Semantic prior
only} removes the residual ($r = 0$), so that
$\mathbf{e}_c = P_{\mathrm{sem}}(\mathbf{z}_c)$, and \emph{Residual
embeddings only} removes the semantic pathway, so that
$\mathbf{e}_c = \mathbf{r}_c$. \emph{Frozen isometry + residual} fixes
$P_{\mathrm{sem}} = \alpha Q$ as discussed in Section~\ref{sec:event_representation}, while
retaining the residual, and \emph{Frozen isometry only} also removes the
residual; both retain the BioLORD vectors rather than replacing them with
random vectors. In every configuration, the modality embedding, temporal
biases, output bias, and logit scale remain learned, so ``only'' in the
configuration names refers to the event-code pathways. On OLMo 1B, we additionally run a within-modality permutation control, which
reassigns BioLORD vectors among events of the same modality to break their
correspondence to clinical concepts while preserving the vectors and the
architecture (Appendix~\ref{app:permutation}), and a residual-rank sweep over
$r \in \{0, 16, 64, 256, 512\}$ (Appendix~\ref{app:rank-sweep}).

\paragraph{Optimization and metrics.}
EHRAdapt parameters are trained with AdamW for at most six epochs on every backbone, with checkpoints selected by validation cross-entropy (CE) and early stopping. We report test CE globally and within ten deciles of training frequency, from D1 (rarest) to D10 (most frequent). Results are mean $\pm$ SD across three seeds, with ablation effects paired within seeds. Training details and additional metrics are given in Appendices~\ref{app:hyper} and~\ref{app:additional_cpt_test_results}.

\subsection{Biosurveillance tasks and cohorts}
\label{sec:downstream-tasks}
Recognizing reportable infections and syndromic illness from clinical
histories can support biosurveillance. Following the task designs of
\citet{goncalves2025biosurveillance}, we evaluate whether EHR representations
distinguish infectious diagnoses and gastrointestinal or respiratory
syndromes before the index diagnosis. To be clear, these patient-level tasks assess a
component of surveillance, they do not measure outbreak detection or alert
lead time. Appendix~\ref{app:downstream} provides more information including all class-by-year counts, cumulative training counts, and disease-code definitions.

\paragraph{ICD-based syndromic classification.}
Syndromic surveillance usually relies on chief complaints \citep{henning2004syndromic}; following \citet{goncalves2025biosurveillance}, we instead define syndromes by sets of nonspecific ICD codes. The four classes are gastrointestinal (GI) syndrome, respiratory syndrome, other known disease, and matched control, with approximately 118 thousand examples. We report the four-class macro average and the GI and respiratory one-versus-rest scores.

\paragraph{Reportable infectious-disease classification.}
Classes are defined by ICD codes for specific reportable infections. The 14 classes comprise seven named higher-frequency diseases, six clinically grouped rare-disease classes, and matched control, with approximately 19.1 thousand examples. We report All 14, Diseases 13 (excluding control), Frequent 7, and Rare 6, the principal view when disease labels are scarce. All views average class metrics from the same 14-way model rather than separately trained ones.


For each example, models receive the patient's clinical events from the two
months preceding the index date, excluding the index date itself, and predict
the example's class, a diagnosis or control label. We use an expanding temporal evaluation: models are
trained on 2019 and tested on 2020, trained on 2019--2020 and tested on 2021,
and trained on 2019--2021 and tested on 2022.

\subsection{Downstream models and baselines}
\label{sec:downstream-models}

We train $\ell_2$-regularized logistic regression probes on each CPT model's hidden state at the last event before the index date. The regularization strength $C \in \{0.1, 1, 10\}$ is chosen per representation, task, and training window using patient-grouped validation, and the chosen probe is refit on the full window. To address class imbalance, we weighted each class by the inverse of its effective number of samples, computed from class frequencies in the training set, using $\beta=0.999$ \citep{cui2019classbalanced}. The same protocol applies to the \textit{Full model}, its ablations, and the baselines below.

\paragraph{Baselines.}
Because EHRAdapt is model-agnostic, our goal is not to find the best backbone or the best task-specific predictor, but to measure the quality of the learned representations for both rare and frequent diseases. We therefore keep all representations frozen, use the same probe for all, and choose baselines that each isolate one aspect of EHRAdapt:
\begin{itemize}
\item \textit{Native text:} the same backbones read the patient history written as text, and the probe uses the hidden state of the last token. This holds the backbone fixed and changes only how events enter it.
\item \textit{BioLORD:} each event is represented by its frozen BioLORD embedding, and these are averaged into a patient vector, either uniformly (mean) or with weights $2^{-d_i}$, where $d_i$ is the event's age in days (decay, a one-day half-life). This tests whether clinical knowledge alone, without sequence modeling or EHR training, is sufficient.
\item \textit{Next-event prediction (NEP):} we implement the NEP training strategy \citep{chen2025buildingehrfoundationmodel} to adapt the same backbones to the CPT training split serialized as text.
  Each instruction-response example contains a chronological patient history and asks the model to predict one subsequent event, with targets reweighted so that each modality (e.g., diagnosis, laboratory test, or medication) is sampled approximately in proportion to the square root of its frequency.
  As in the original, we train rank-16 LoRA adapters for 10,000 optimizer steps with a global batch of 512, using a 4,096-token input limit, and then freeze the adapted model.
  The patient representation is the mean of the final-layer hidden states over nonpadding tokens in the serialized history.
  Appendix~\ref{app:nep-ehradapt-training} details its training and compares it with that of EHRAdapt.
  This tests whether standard next-event language-model adaptation produces transferable patient representations when the backbone, EHR data, and downstream probe are held fixed.
  
\item \textit{Random forest:} trained on counts of clinical events in the same input window, with the number of trees tuned over $\{100, 200, \ldots, 500\}$. Such count-based models are strong baselines for structured EHR prediction \citep{wornow2023ehrshot,Brown_2025}.
\end{itemize}
Medical language models such as GatorTronGPT \citep{Peng2023-sm} could serve as EHRAdapt backbones, and any of these representations could be further fine-tuned for each task; we leave both backbone comparison and task-specific fine-tuning outside our scope.

\paragraph{Access to pretraining data.}
All methods train their probe or classifier on the same labeled downstream windows and differ only in whether they also use the unlabeled 2010--2018 CPT corpus,\footnote{The CPT records carry no downstream labels but contain the diagnosis codes used to define them.} consistent with standard evaluations of pretrained EHR representations against count-based models \citep{wornow2023ehrshot}. Native text, BioLORD, and the random forest use no CPT records, although the backbones and BioLORD carry knowledge from their own general-domain and ontology pretraining. In contrast, EHRAdapt and NEP are both adapted on the CPT training split, so NEP provides a data-matched comparison: both methods train on essentially all 3.95 million training patients, keep the backbone frozen, and have a comparable number of trainable parameters (6.0--8.4 million for NEP and 6.8--8.5 million for EHRAdapt; Table~\ref{tab:nep-ehradapt-training}). However, the two methods use these records differently. NEP serializes each history as text, at about 27 tokens per event, and supervises one sampled event per history, so its 10,000 steps cover the training chunks 1.15 times and yield 5.12 million supervised targets; EHRAdapt represents each event as a single token and supervises every position, yielding 1.55--3.41 billion targets. NEP nonetheless processed more backbone tokens and GPU-hours than EHRAdapt on every backbone, so its lower supervision reflects the cost of text serialization rather than a smaller training budget. Furthermore, the five EHRAdapt configurations share identical CPT data, so differences among them reflect the event representation rather than data access (Table~\ref{tab:downstream-components}). The CPT corpus ends in 2018, before every downstream index date.



\section{Results}
\label{sec:results}

This section reports next-event prediction results from continued pretraining, downstream classification performance of EHRAdapt and baseline models, and ablations of EHRAdapt's components.

\subsection{Continued-pretraining performance}

Table~\ref{tab:aggregate} shows that EHRAdapt's \textit{Full model} achieves the lowest global test CE among the five configurations on all three backbones. This ordering holds for all paired ablation vs \textit{Full model} comparisons on metrics evaluated. Learning the semantic projection also matters: fixing it (\textit{Frozen isometry}) worsens the global CE and on every data stratum compared with its learned counterpart. 
See Appendix~\ref{app:additional_cpt_test_results} for all accuracy results and performance curves for all ablations on each data stratum.

\begin{table}[htb]
\centering\small
\setlength{\tabcolsep}{3pt}
\caption{Global CPT test CE (nats, lower is better), mean $\pm$ sample SD across three seeds.}
\label{tab:aggregate}
\begin{tabular}{@{}lrrr@{}}
\toprule
Configuration & OLMo2-1B & Llama3.2-1B & OLMo2-7B\\
\midrule
Full model & $\mathbf{2.847}\pm0.006$ & $\mathbf{2.806}\pm0.011$ & $\mathbf{2.799}\pm0.003$\\
Semantic prior only & $3.217\pm0.019$ & $3.129\pm0.006$ & $3.090\pm0.008$\\
Residual embeddings only & $2.959\pm0.012$ & $2.937\pm0.014$ & $2.976\pm0.050$\\
Frozen isometry + residual & $3.158\pm0.027$ & $2.974\pm0.015$ & $3.057\pm0.001$\\
Frozen isometry only & $4.689\pm0.005$ & $4.500\pm0.017$ & $4.513\pm0.014$\\
\bottomrule
\end{tabular}
\end{table}

\begin{figure}[htb]
\centering
\includegraphics[width=0.9\linewidth]{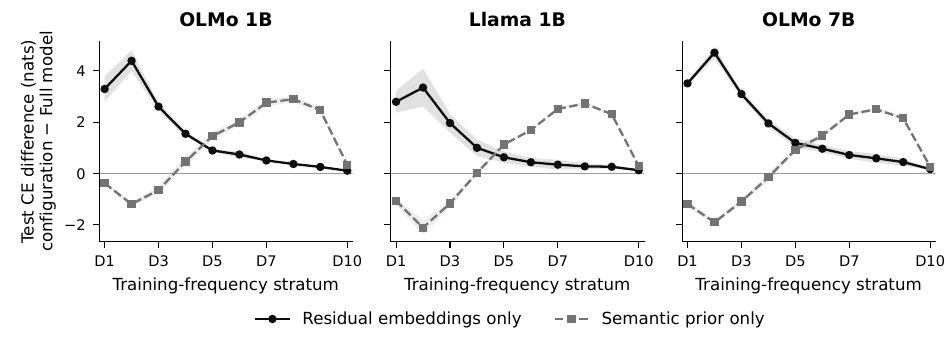}
\caption{Ablation's test CE minus the \textit{Full model}'s for each training frequency: positive values mean the ablation is worse. \textit{Residual embeddings only} removes the semantic pathway, and \textit{Semantic prior only} removes the residual. D1 is rarest and D10 most frequent. Curves show mean and $\pm$1 sample SD over three seeds.}
\label{fig:decile}
\end{figure}

Figure~\ref{fig:decile} shows how test CE changes for each data stratum when each pathway is removed. Removing the semantic pathway (\textit{Residual embeddings only}) hurts rare events most, and the penalty shrinks steadily as events become more frequent. Removing the residual (\textit{Semantic prior only}) shows the opposite pattern: it improves the rarest events (D1--D3) but hurts events of medium to high frequency. The two curves cross around D5 on all three backbones, and both approach zero in D10, suggesting that abundant data make either pathway nearly sufficient on its own. As aggregate CE is dominated by frequent events, these large effects on rare events have little effect on the total loss.


We also run a permutation control on OLMo2-1B that shuffles BioLORD embeddings within each modality. This keeps the set of vectors and their geometry unchanged but breaks the link between each event and its clinical meaning. The permuted model has higher test CE than the \textit{Full model} globally and in every stratum, indicating that the semantic pathway helps because each event receives its correct clinical description, not merely because of how the vectors are distributed (Appendix~\ref{app:permutation}).

\subsection{Downstream predictive performance}
\label{sec:downstream}

Table~\ref{tab:downstream-performance} compares the \textit{Full model} with the baselines. Overall, the \textit{Full model} outperforms every baseline in Table~\ref{tab:downstream-performance} on every reported view and on all three backbones. The strongest baseline depends on the task: the tuned random forest leads the baselines on every infectious-disease view and on GI, whereas NEP leads on the four-class syndromic average and on respiratory. Relative to the strongest baseline, the \textit{Full model} improves Rare-6 AUPRC from 0.190 to 0.233--0.247 and four-class syndromic AUPRC from 0.732 to 0.791--0.801. The margin is smallest for GI (0.681--0.686 versus 0.667 for the random forest), and for respiratory it is 0.730--0.754 versus 0.699 for NEP.

NEP, which adapts the same backbones on the same CPT corpus, shows a task-dependent pattern. On the syndromic task, it is the strongest baseline for the four-class average (0.715--0.732) and for respiratory (0.672--0.699), well above native text (0.530--0.572 for respiratory), but it trails both the random forest and BioLORD on GI. In contrast, on the infectious-disease task NEP falls below native text on 11 of 12 backbone--view pairs, with Rare-6 AUPRC of 0.088--0.117 (Table~\ref{tab:downstream-performance}). Next-event adaptation of text-serialized records therefore did not transfer to rare-disease discrimination in our setting, and on Llama 1B it varied substantially across seeds on the syndromic task (SD 0.048 for the four-class average and 0.099 for GI). We hypothesize that mean pooling over long serialized histories dilutes the few diagnostic events that distinguish rare infections, whereas syndromic classes reflect broader patterns among common events; this warrants further investigation. The \textit{Full model} exceeds NEP on the same backbone by 0.127--0.162 on All 14, 0.130--0.151 on Rare 6, and 0.069--0.082 on the four-class syndromic average. Because NEP is adapted on the same patients with comparable trainable parameters and more compute, these gaps indicate that EHRAdapt's downstream advantage is not explained by access to pretraining data or by training budget.

The \textit{Full model} outperforms both downstream task averages among the five EHRAdapt configurations, consistent with its global CPT ranking. However, better CPT likelihood does not always transfer: removing the residual improves rare-event likelihood but reduces Rare-6 AUPRC by 0.9--2.1 percentage points. Removing the semantic branch costs 3.6--5.1 Rare-6 points, versus 1.3--2.5 for Frequent 7, while freezing the projection with the residual retained costs 3.2--4.0 Rare-6 points (Table~\ref{tab:downstream-components}). These Rare-6 effects persist across backbones, supporting the contributions of both shared projection learning and event-specific adaptation to rare-disease discrimination.



    \begin{table}[tb]
    \centering\small
    \setlength{\tabcolsep}{3pt}
    \caption{Downstream AUPRC. Multiclass tasks use macro-AUPRC. Scores are mean $\pm$ sample SD over seeds (each seed averaged over 2020--2022).}
    \label{tab:downstream-performance}
    \begin{tabular}{@{}lrrrr@{}}
    \toprule
    Model & All 14 & Diseases 13 & Frequent 7 & Rare 6\\
    \midrule
    BioLORD (mean) & $0.297$ & $0.254$ & $0.345$ & $0.148$\\
    BioLORD (decay) & $0.308$ & $0.266$ & $0.356$ & $0.162$\\
    OLMo 1B, native text & $0.266$ & $0.221$ & $0.303$ & $0.126$\\
    Llama 1B, native text & $0.261$ & $0.215$ & $0.302$ & $0.113$\\
    OLMo 7B, native text & $0.255$ & $0.209$ & $0.290$ & $0.114$\\
    RF, tuned (count-based) & $0.330$ & $0.288$ & $0.373$ & $0.190$\\
    NEP - OLMo 1B & $0.230\pm0.011$ & $0.183\pm0.012$ & $0.256\pm0.013$ & $0.098\pm0.010$\\
    NEP - Llama 1B & $0.253\pm0.004$ & $0.207\pm0.004$ & $0.285\pm0.005$ & $0.117\pm0.004$\\
    NEP - OLMo 7B & $0.213\pm0.004$ & $0.165\pm0.005$ & $0.230\pm0.004$ & $0.088\pm0.007$\\
    \midrule
    EHRAdapt - OLMo 1B & $0.366\pm0.004$ & $0.326\pm0.005$ & $0.405\pm0.002$ & $0.233\pm0.010$\\
    EHRAdapt - Llama 1B & $\boldsymbol{0.380\pm0.002}$ & $\boldsymbol{0.341\pm0.002}$ & $\boldsymbol{0.421\pm0.002}$ & $\boldsymbol{0.247\pm0.004}$\\
    EHRAdapt - OLMo 7B & $0.375\pm0.006$ & $0.335\pm0.007$ & $0.418\pm0.005$ & $0.239\pm0.010$\\
    \midrule
    Model & All 4 & GI & Respiratory & \\
    \midrule
    BioLORD (mean) & $0.651$ & $0.590$ & $0.453$ & \\
    BioLORD (decay) & $0.669$ & $0.605$ & $0.491$ & \\
    OLMo 1B, native text & $0.690$ & $0.586$ & $0.566$ & \\
    Llama 1B, native text & $0.670$ & $0.569$ & $0.530$ & \\
    OLMo 7B, native text & $0.691$ & $0.579$ & $0.572$ & \\
    RF, tuned (count-based) & $0.700$ & $0.667$ & $0.486$ & \\
    NEP - OLMo 1B & $0.721\pm0.007$ & $0.569\pm0.015$ & $0.672\pm0.011$ & \\
    NEP - Llama 1B & $0.715\pm0.048$ & $0.547\pm0.099$ & $0.681\pm0.035$ & \\
    NEP - OLMo 7B & $0.732\pm0.010$ & $0.578\pm0.019$ & $0.699\pm0.009$ & \\
    \midrule
    EHRAdapt - OLMo 1B & $0.791\pm0.005$ & $0.681\pm0.005$ & $0.730\pm0.009$ & \\
    EHRAdapt - Llama 1B & $0.797\pm0.003$ & $\boldsymbol{0.686\pm0.005}$ & $0.741\pm0.004$ & \\
    EHRAdapt - OLMo 7B & $\boldsymbol{0.801\pm0.001}$ & $0.682\pm0.001$ & $\boldsymbol{0.754\pm0.002}$ & \\
    \bottomrule
    \end{tabular}
  \end{table}

\begin{table}[htb]
\centering\small
\setlength{\tabcolsep}{3pt}
\caption{\textit{Full model} improvement over ablations (positive favors \textit{Full model}): reduction in CPT test cross-entropy for $\Delta$CE (nats) and gain in AUPRC for downstream columns (percentage points), paired by seed and year. GI/Resp. averages the two class AUPRCs. Full tables in Appendix~\ref{app:downstream-results}.}
\label{tab:downstream-components}
\begin{tabular}{@{}lrrrr@{}}
\toprule
Configuration & $\Delta$CE & Freq. 7 & Rare 6 & GI/Resp.\\
\midrule
\multicolumn{5}{l}{\textit{OLMo 1B}}\\
Semantic prior only & $0.371$ & $+0.38$ & $+1.91$ & $+0.30$\\
Residual embeddings only & $0.112$ & $+1.30$ & $+3.60$ & $+0.66$\\
Frozen isometry + residual & $0.312$ & $+2.69$ & $+3.30$ & $+6.69$\\
Frozen isometry only & $1.842$ & $+5.95$ & $+8.48$ & $+6.78$\\
\midrule
\multicolumn{5}{l}{\textit{Llama 1B}}\\
Semantic prior only & $0.323$ & $+0.54$ & $+0.86$ & $+1.82$\\
Residual embeddings only & $0.131$ & $+2.16$ & $+4.16$ & $+1.95$\\
Frozen isometry + residual & $0.168$ & $+1.43$ & $+3.95$ & $+1.66$\\
Frozen isometry only & $1.694$ & $+4.43$ & $+7.64$ & $+3.15$\\
\midrule
\multicolumn{5}{l}{\textit{OLMo 7B}}\\
Semantic prior only & $0.291$ & $+1.06$ & $+2.08$ & $+0.39$\\
Residual embeddings only & $0.177$ & $+2.47$ & $+5.10$ & $+0.72$\\
Frozen isometry + residual & $0.258$ & $+1.47$ & $+3.21$ & $+1.61$\\
Frozen isometry only & $1.714$ & $+6.52$ & $+9.17$ & $+4.33$\\
\bottomrule
\end{tabular}
\end{table}

\subsection{Representation Geometry}
\label{sec:representation}

\begin{figure}[htb]
    \centering
    \includegraphics[width=0.95\linewidth]{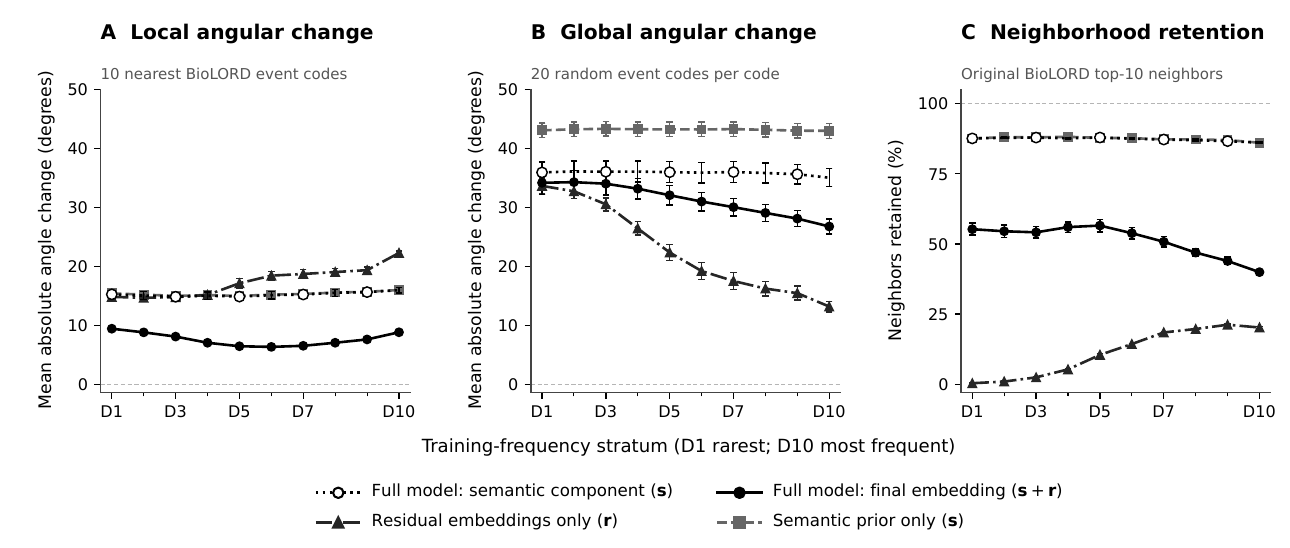}
    \caption{\textbf{BioLORD geometry preservation after CPT (OLMo2-1B).} Angular changes relative to frozen, mean-centered BioLORD are measured between each event-code embedding and those of (A) its 10 original nearest neighbors or (B) 20 randomly sampled codes. (C) Percentage of original top-10 neighbors retained. Curves show the \textit{Full model}’s semantic component and final embedding, alongside the final embeddings of separately trained ablations. Each frequency stratum uses 1,000 shared event codes; measurements are averaged equally over codes. Points and error bars show the mean and sample SD across three training seeds.}
    \label{fig:biolord-preservation}
\end{figure}

We examine how EHRAdapt retains the semantic structure supplied by BioLORD while adapting event representations to EHR data. Figure~\ref{fig:biolord-preservation} shows how local neighborhood and angular information is preserved at each data stratum, using the OLMo2-1B backbone. The learned semantic projection retains most of BioLORD's nearest neighbors (C) but substantially changes the angles between both neighboring and randomly paired codes (A, B). Thus, learning $P_{\mathrm{sem}}$ largely preserves which codes are most similar while changing their similarity values.

Adding the residual further reorganizes local neighborhoods, particularly for frequent events. However, it reduces average angular distortion relative to BioLORD in both local and global comparisons. Preserving similarity values and preserving neighbor rankings thus capture different aspects of the learned representations.

\section{Conclusion}

We introduced EHRAdapt, an adapter that maps structured EHR events into frozen pretrained LLMs via a semantic prior and learned residual. Across three backbones, the full model had the best test likelihood. The semantic prior benefited rare events most, while the residual improved overall likelihood but degraded next-event prediction for the rarest events. Both components improved rare-disease classification, showing the need to evaluate across event frequencies and downstream tasks.

Continued pretraining on longitudinal EHR data produced representations that transferred effectively to downstream tasks and outperformed the evaluated baselines, including text-based inputs to the same LLMs and LLMs adapted on the same EHR corpus through next-event prediction. Limitations include the exclusion of laboratory and vital-sign values and demographic information, and evaluation within a single health system. Future work should incorporate these features and assess generalizability across healthcare systems.





\section*{Acknowledgments}

This work was performed under the auspices of the U.S. Department of Energy by Lawrence
Livermore National Laboratory under Contract DE-AC52-07NA27344. Release number: LLNL-CONF-2024786.

\bibliographystyle{plainnat}
\bibliography{references}

\clearpage
\appendix
\section*{Supplementary information}

\section{Data and preprocessing}
\label{app:data}

\begin{figure}[htb]
    \centering
    \includegraphics[width=0.8\linewidth]{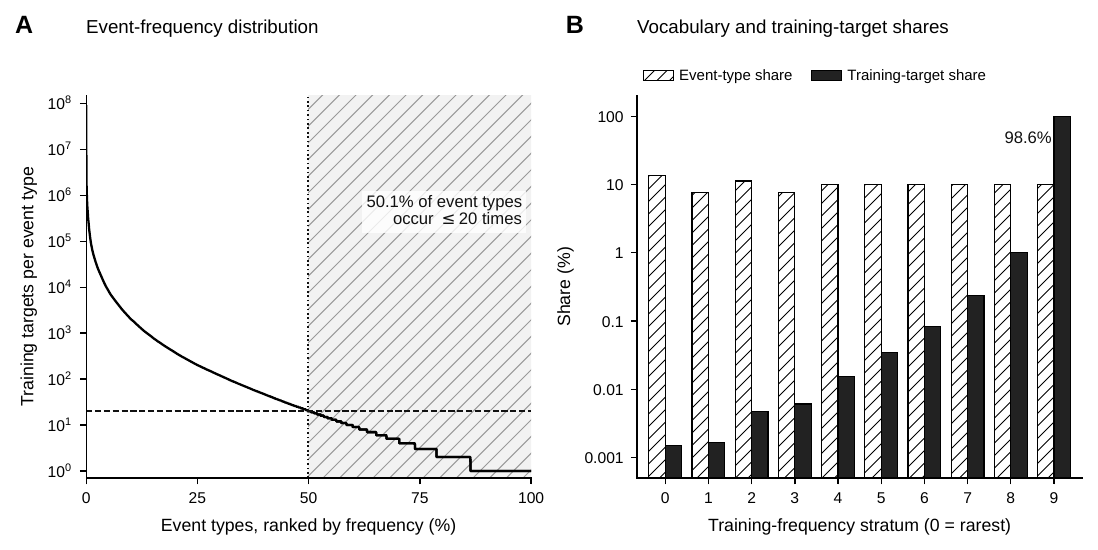}
    \caption{Clinical event frequencies in the KPNC continued-pretraining corpus.
Counts refer to the training split and exclude reserved tokens and
end-of-sequence targets.}
    \label{fig:kp-long-tail}
\end{figure}

Table~\ref{tab:data} summarizes the continued-pretraining cohort and its
patient-level splits, and Figure~\ref{fig:kp-long-tail} shows the frequency
distribution of clinical events in the training split. Frequency strata are
derived from training-split counts. The decile boundaries are $1$, $1$, $2$,
$5$, $9$, $20$, $47$, $120$, $379$, $2{,}050$, and approximately $87.6$M
occurrences. The repeated boundary at one reflects tied singleton frequencies,
so D1--D10 are not equal-sized bins, and all reported metrics use exact bin
memberships. The vocabulary has 78,384 clinical event types and four reserved
tokens, for $|\mathcal{V}|$ = 78{,}388.

\begin{table}[htbp]
\centering\small
\caption{Continued-pretraining cohort and splits. Patients with at least one clinical event. Chunking retains all 732M clinical events. Event types by modality:
diagnosis 44,079; procedure 28,401; laboratory test 2,853; chief complaint
2,174; medication 817; encounter type 59; vital signs 1.}
\label{tab:data}
\begin{tabular}{@{}lrrrrr@{}}
\toprule
Split & Fraction & Patients & Chunks & Clinical events \\
\midrule
Train      & 98\% & 3.95M  & 4.46M  & 717M \\
Validation & 1\%  & 40,197 & 45,484 & 7.32M \\
Test       & 1\%  & 39,734 & 45,144 & 7.32M \\
\bottomrule
\end{tabular}
\end{table}

\section{Backbones and parameter counts}
\label{app:sec_backbone_models}

Table~\ref{tab:backbones} lists the three frozen backbones. The mean norm of
each backbone's pretrained token embeddings sets the initialization scale of
the semantic projection (Appendix~\ref{app:hyper}).
Table~\ref{tab:param-counts} gives parameter counts: EHRAdapt trains
0.1--0.6\% of the parameters of the complete model.

\begin{table}[htbp]
\centering\footnotesize
\setlength{\tabcolsep}{5pt}
\caption{Frozen backbones. $H$: hidden size; Emb.\ norm: mean norm of the pretrained token embeddings, which sets the initialization scale of $P_{\mathrm{sem}}$.}
\label{tab:backbones}
\begin{tabular}{@{}llrrrrr@{}}
\toprule
Backbone & Checkpoint & $H$ & Layers & Heads & KV heads & Emb.\ norm \\
\midrule
OLMo2-1B & \texttt{allenai/OLMo-2-0425-1B} & 2,048 & 16 & 16 & 16 & 10.724 \\
Llama3.2-1B & \texttt{meta-llama/Llama-3.2-1B} & 2,048 & 16 & 32 & 8 & 0.987 \\
OLMo2-7B & \texttt{allenai/OLMo-2-1124-7B} & 4,096 & 32 & 32 & 32 & 7.423 \\
\bottomrule
\end{tabular}
\end{table}

\begin{table}[!ht]
\centering
\footnotesize
\setlength{\tabcolsep}{5pt}
\caption{Parameter counts for the full EHRAdapt configuration (78,388-event vocabulary, linear semantic projection, rank-64 projected residual, tied prediction head, frozen backbone). The original OLMo2 text models have separate language-model heads, which EHRAdapt replaces; Llama3.2 ties its head to its input embeddings, so its two backbone counts are identical. In panel B, $V$ is the vocabulary size, $r$ the residual rank, $H$ the hidden size, $L$ and $A$ the numbers of layers and attention heads, and $K$ the number of elapsed-time buckets; 768 is the BioLORD embedding dimension.}
\label{tab:param-counts}
\begin{tabular}{@{}llrrr@{}}
\toprule
 & Size & OLMo2-1B & Llama3.2-1B & OLMo2-7B \\
\midrule
\multicolumn{5}{@{}l}{\textit{A. Totals}} \\
Original text LM, including LM head & & 1,484,916,736 & 1,235,814,400 & 7,298,617,344 \\
Frozen backbone used by EHRAdapt & & 1,279,395,840 & 1,235,814,400 & 6,887,575,552 \\
Trainable EHRAdapt parameters & & 6,821,685 & 6,825,781 & 8,556,341 \\
Complete EHRAdapt model & & 1,286,217,525 & 1,242,640,181 & 6,896,131,893 \\
Trainable fraction & & 0.53\% & 0.55\% & 0.12\% \\
\midrule
\multicolumn{5}{@{}l}{\textit{B. Trainable parameters by component}} \\
Semantic projection & $768H + H$ & 1,574,912 & 1,574,912 & 3,149,824 \\
Event residual table & $Vr$ & 5,016,832 & 5,016,832 & 5,016,832 \\
Residual projection & $rH$ & 131,072 & 131,072 & 262,144 \\
Modality embeddings & $8H$ & 16,384 & 16,384 & 32,768 \\
Temporal attention biases & $LAK$ & 4,096 & 8,192 & 16,384 \\
Prediction bias and logit scale & $V + 1$ & 78,389 & 78,389 & 78,389 \\
\midrule
Total trainable & & 6,821,685 & 6,825,781 & 8,556,341 \\
\bottomrule
\end{tabular}
\end{table}

\section{Training configuration}
\label{app:hyper}

Table~\ref{tab:training-config} lists the training settings. The backbone
receives no gradient updates, so only EHRAdapt parameters are optimized.
Within each backbone, all configurations use the same settings. Runs use AMD
MI300A accelerators under ROCm.

\paragraph{Initialization.}
The semantic projection $P_{\mathrm{sem}}$ is a learned linear map with a
scaled semi-orthogonal initialization. The scale is set so that projected
semantic vectors match the mean norm of the backbone's pretrained token
embeddings (Table~\ref{tab:backbones}), so event inputs enter the transformer
at a magnitude comparable to its native token inputs. This norm serves only as
an initialization target: $P_{\mathrm{sem}}$, $\mathbf{A}$, $\mathbf{B}$, and
$\mathbf{e}_{\mathrm{mod}}$ are all trained during CPT. The output bias
$\beta_c$ is initialized to $\log p(c)$ from training-split counts, which
lowers the initial CE by 4.899 nats relative to a uniform initialization
($\log V = 11.270$ versus a unigram entropy of 6.371). The logit scale $s$ is
calibrated at initialization by one-dimensional minimization on a
training-only calibration subset, separately for each configuration, because
different initializations produce differently scaled logits.

\paragraph{BioLORD's embedding processing.}
Each event's BioLORD embedding is obtained by mean pooling the token embeddings of its description and L2-normalizing the result. These embeddings are then centered by subtracting their mean over all clinical event types, without whitening.

\begin{table}[htbp]
\centering\footnotesize
\setlength{\tabcolsep}{5pt}
\caption{EHRAdapt training configuration. Entries spanning all three backbone columns
  are shared across backbones.}
\label{tab:training-config}
\begin{tabular}{@{}llll@{}}
\toprule
Setting & OLMo2 1B & Llama3.2 1B & OLMo2 7B \\
\midrule
Backbone optimization & \multicolumn{3}{l}{Frozen} \\
Optimizer & \multicolumn{3}{l}{AdamW} \\
Precision; attention & \multicolumn{3}{l}{BF16; \texttt{sdpa}} \\
Effective batch & \multicolumn{3}{l}{128 chunks (no gradient accumulation)} \\
Learning rate: output/temporal &
  \multicolumn{3}{l}{$1\mathrm{e}{-4}$ for vocabulary bias, logit scale, and temporal-attention bias} \\
  Learning rate: event encoder &
  \multicolumn{3}{l}{$3\mathrm{e}{-4}$ for semantic projection, residual parameters, and modality embeddings} \\
  Weight decay &
  \multicolumn{3}{l}{0.1 on semantic and residual projection weights; 0 otherwise} \\
Learning-rate schedule & \multicolumn{3}{l}{1\% warmup, constant, final 10\% decay to $0.1\times$} \\
Gradient clipping & \multicolumn{3}{l}{Global norm 1.0} \\
Validation interval & \multicolumn{3}{l}{5,000 steps} \\
Checkpoint selection & \multicolumn{3}{l}{Target-weighted global validation CE} \\
Early stopping & \multicolumn{3}{l}{Patience 3; $\Delta_{\min}=0.005$ nats} \\
Validation CE ceiling & \multicolumn{3}{l}{6.371 nats (unigram entropy)} \\
Length-bucket boundaries & \multicolumn{3}{l}{32, 64, 128, 256, 384, 512, 527} \\
Training seeds; split seed & \multicolumn{3}{l}{101, 202, 303; 17} \\
Maximum valid-target budget &
    \multicolumn{3}{l}{$4.326\times10^9$} \\
Maximum data passes & \multicolumn{3}{l}{6} \\
\midrule
Scheduler horizon (steps) & 213,449 & 213,449 & 209,196 \\
Steps per epoch & $\sim$35.6K & $\sim$35.6K & $\sim$34.9K \\
Nodes / GPU ranks & 4 / 16 & 4 / 16 & 16 / 64 \\
Micro-batch per rank & 8 & 8 & 2 \\
  
\bottomrule
\end{tabular}
\end{table}


\section{Downstream cohorts}
\label{app:downstream}

\subsection{Class counts}
\label{app:downstream-counts}
Tables~\ref{tab:downstream-ili-syndromic-counts}
and~\ref{tab:downstream-infectious7-rare6-counts} give class counts by year,
and Table~\ref{tab:downstream-evaluated-counts} gives the training and test
sizes of each evaluation window. All positive cases are retained. Controls are
capped at one demographically matched control per case, so the control-to-case
ratio can fall below 1:1. Annual columns are disjoint sets of examples, and
each training window accumulates all earlier years.

\begin{table}[htb]
\centering\footnotesize
\setlength{\tabcolsep}{3pt}
\caption{Syndromic classification case counts per year. Each test year uses all preceding years for training.}
\label{tab:downstream-ili-syndromic-counts}
\begin{tabular}{@{}>{\raggedright\arraybackslash}p{0.41\linewidth}rrrrr@{}}
\toprule
Class & 2019 & 2020 & 2021 & 2022 & Total\\
\midrule
Gastrointestinal & 1,420 & 742 & 808 & 1,150 & 4,120\\
Respiratory & 9,474 & 6,943 & 3,765 & 6,899 & 27,081\\
Known & 10,894 & 7,685 & 4,573 & 8,049 & 31,201\\
Control & 19,447 & 13,524 & 8,261 & 14,376 & 55,608\\
\midrule
All positive cases & 21,788 & 15,370 & 9,146 & 16,098 & 62,402\\
\textbf{Total} & \textbf{41,235} & \textbf{28,894} & \textbf{17,407} & \textbf{30,474} & \textbf{118,010}\\
\bottomrule
\end{tabular}
\end{table}

\begin{table}[htbp]
\centering\footnotesize
\setlength{\tabcolsep}{3pt}
\caption{Infectious-disease classification case counts per year. Each test year uses all preceding years for training. Rare-6 diseases are identified by $^{\dagger}$.}
\label{tab:downstream-infectious7-rare6-counts}
\begin{tabular}{@{}>{\raggedright\arraybackslash}p{0.41\linewidth}rrrrr@{}}
\toprule
Class & 2019 & 2020 & 2021 & 2022 & Total\\
\midrule
Campylobacteriosis & 825 & 448 & 479 & 634 & 2,386\\
Salmonellosis (nontyphoidal) & 590 & 357 & 298 & 481 & 1,726\\
Chickenpox (varicella) & 386 & 255 & 178 & 218 & 1,037\\
Shigellosis & 419 & 180 & 246 & 360 & 1,205\\
Pertussis & 486 & 97 & 28 & 30 & 641\\
Invasive \textit{H. influenzae} & 190 & 118 & 63 & 107 & 478\\
Viral meningitis & 126 & 93 & 103 & 93 & 415\\
$^{\dagger}$Rare vector/travel systemic & 142 & 66 & 60 & 80 & 348\\
$^{\dagger}$Rare enteric diarrheal & 99 & 81 & 70 & 71 & 321\\
$^{\dagger}$Rare vaccine preventable respiratory/neurologic & 129 & 65 & 59 & 45 & 298\\
$^{\dagger}$Rare environmental food/water & 79 & 65 & 73 & 67 & 284\\
$^{\dagger}$Hepatitis A & 80 & 71 & 51 & 52 & 254\\
$^{\dagger}$Rare zoonotic/tick borne high consequence & 59 & 51 & 59 & 49 & 218\\
Control & 3,582 & 1,903 & 1,756 & 2,277 & 9,518\\
\midrule
All positive cases & 3,610 & 1,947 & 1,767 & 2,287 & 9,611\\
\textbf{Total} & \textbf{7,192} & \textbf{3,850} & \textbf{3,523} & \textbf{4,564} & \textbf{19,129}\\
\bottomrule
\end{tabular}
\end{table}

\begin{table}[htbp]
\centering\small
\caption{Full-refit training and held-out annual test counts for syndromic and infectious disease tasks.}
\label{tab:downstream-evaluated-counts}
\begin{tabular}{lrrrrrr}
\toprule
& \multicolumn{2}{c}{Test 2020} & \multicolumn{2}{c}{Test 2021} & \multicolumn{2}{c}{Test 2022}\\
Task & Train & Test & Train & Test & Train & Test\\
\midrule
Syndromic & 41,235 & 28,894 & 70,129 & 17,407 & 87,536 & 30,474\\
Infectious & 7,192 & 3,850 & 11,042 & 3,523 & 14,565 & 4,564\\
\bottomrule
\end{tabular}
\end{table}

\subsection{Class definitions}
\label{app:disease_class_codes}
Table~\ref{tab:class-definitions} lists the ICD codes that define each class.
In syndromic classification, the Known class contains patients with a case
diagnosis outside the two syndromic code lists, sampled at random with a fixed
seed. Control patients are matched to patients in the disease classes by
demographics and geographic location. The sizes of the Known and Control
classes were set to keep the classes balanced.

\begin{table}[!ht]
\centering
\scriptsize
\setlength{\tabcolsep}{3pt}
\renewcommand{\arraystretch}{1.1}
\caption{ICD codes defining the classes in the two downstream tasks. Ranges are inclusive.}
\label{tab:class-definitions}
\begin{tabularx}{\linewidth}{@{}
>{\raggedright\arraybackslash}p{0.19\linewidth}
>{\raggedright\arraybackslash\hsize=0.8\hsize}X
>{\raggedright\arraybackslash\hsize=1.2\hsize}X@{}}
\toprule
Class & ICD-9-CM & ICD-10-CM \\
\midrule
\multicolumn{3}{@{}l}{\textit{Syndromic classification}} \\
Gastrointestinal &
-- &
A02.9, A03.0--A03.3, A03.8--A03.9, A04.5, A05.1, A05.3, A05.5, A32.11, A32.12, A32.7, A32.89, A32.9, B96.21 \\
Respiratory &
460, 461.9, 465.8, 465.9, 466.0, 486, 487.0, 487.1, 487.8, 488.01, 488.02, 488.09, 488.11, 488.12, 488.81, 488.82, 488.89, 490 &
J00, J01.90, J06.9, J09.X1--J09.X3, J09.X9, J10.00, J10.01, J10.08, J10.1, J10.2, J10.81, J10.89, J11.00, J11.08, J11.1, J11.2, J11.89, J12.89, J12.9, J18.1, J18.8, J18.9, J20.9, J40 \\
Known &
\multicolumn{2}{l@{}}{Any case diagnosis code outside the two syndromic code lists} \\
Control &
\multicolumn{2}{l@{}}{Matched patients (same demographics and geographic location) with none of the codes above} \\
\midrule
\multicolumn{3}{@{}l}{\textit{Infectious-disease classification}} \\
Campylobacteriosis &
008.43 &
A04.5 \\
Salmonellosis (nontyphoidal) &
003.0, 003.1, 003.20, 003.23, 003.29, 003.9 &
A02.0, A02.1, A02.20, A02.22--A02.25, A02.29, A02.8, A02.9 \\
Chickenpox (varicella) &
052.0--052.2, 052.7--052.9 &
B01.0, B01.11, B01.2, B01.81, B01.89, B01.9 \\
Shigellosis &
004.0--004.3, 004.8--004.9 &
A03.0--A03.3, A03.8--A03.9 \\
Pertussis &
033.0, 033.1, 033.9 &
A37.00, A37.01, A37.10, A37.11, A37.80, A37.90, A37.91 \\
Invasive \textit{H.\ influenzae} &
041.5, 482.2 &
A49.2, B96.3, J14, J20.1 \\
Viral meningitis &
047.0, 047.1, 047.8, 047.9 &
A87.0, A87.2, A87.8, A87.9 \\
Rare vector/travel systemic &
061, 084.0, 084.1, 084.4--084.6, 084.9, 066.40--066.42, 066.49, 065.4, 062.0, 062.2, 062.3, 062.9 &
A90, A91, B50.0, B50.8, B50.9, B51.9, B52.0, B52.8, B52.9, B53.8, B54, A92.0, A92.30--A92.32, A92.39, A92.5, A83.0, A83.2, A83.3, A83.5, A83.6, A83.8, A95.1 \\
Rare enteric diarrheal &
283.11, 041.41--041.43, 002.0, 007.4, 005.4, 005.81, 001.0, 001.9, 008.44, 002.1--002.3, 002.9 &
D59.3, D59.30, D59.31, B96.21--B96.23, A01.00, A01.02, A01.03, A01.09, A07.2, A05.3, A05.5, A00.0, A00.9, A04.6, A01.1, A01.3, A01.4 \\
Rare vaccine preventable respiratory/neurologic &
045.00, 045.01, 045.10, 045.90, 046.11, 046.19, 037, 055.79, 055.8, 055.9, 036.0--036.2, 036.42, 036.81, 036.82, 036.89, 036.9, 032.0, 032.3, 032.81, 032.83--032.85, 032.89, 032.9 &
B06.02, B06.89, B06.9, P35.0, Z20.4, A80.30, A80.39, A80.9, B26.0, B26.84, B26.9, A81.00, A81.01, A81.09, A33--A35, B05.2, B05.89, B05.9, A39.0, A39.1, A39.2, A39.4, A39.81, A39.83, A39.9, A36.0, A36.2, A36.3, A36.84, A36.89, A36.9 \\
Rare environmental food/water &
040.89, 482.84, 124, 005.1, 027.0 &
A48.1, A48.2, B75, A05.1, A32.11, A32.12, A32.7, A32.89, A32.9, T61.01XA, T61.11XA, T61.14XA \\
Hepatitis A &
070.0, 070.1 &
B15.0, B15.9 \\
Rare zoonotic/tick borne high consequence &
083.0, 087.0, 087.1, 087.9, 023.0, 023.8, 023.9, 088.82, 065.3, 066.1, 082.9, 083.2, 083.8, 083.9, 071, 082.40, 082.41, 082.49, 021.0, 021.9, 020.2, 020.8, 082.0, 022.0, 022.3, 073.0, 073.8, 073.9 &
A78, A68.0, A68.1, A68.9, A23.0, A23.1, A23.3, A23.8, A23.9, B60.00, B60.02, B60.10, B60.11, B60.13, B60.19, B60.8, A75.2, A75.3, A77.8, A77.9, A79.1, A79.81, A79.89, A79.9, A82.0, A82.1, A82.9, A77.40, A77.41, A77.49, A21.2, A21.7, A21.8, A21.9, A20.0, A20.7, A20.8, A20.9, A77.0, A98.4, A98.5, A22.7, A70 \\
Control &
\multicolumn{2}{l@{}}{Matched patients (same demographics and geographic location) with none of the codes above} \\
\bottomrule
\end{tabularx}
\end{table}

\section{Additional CPT results}
\label{app:additional_cpt_test_results}

\begin{figure}[t]
\centering
\includegraphics[width=\linewidth]{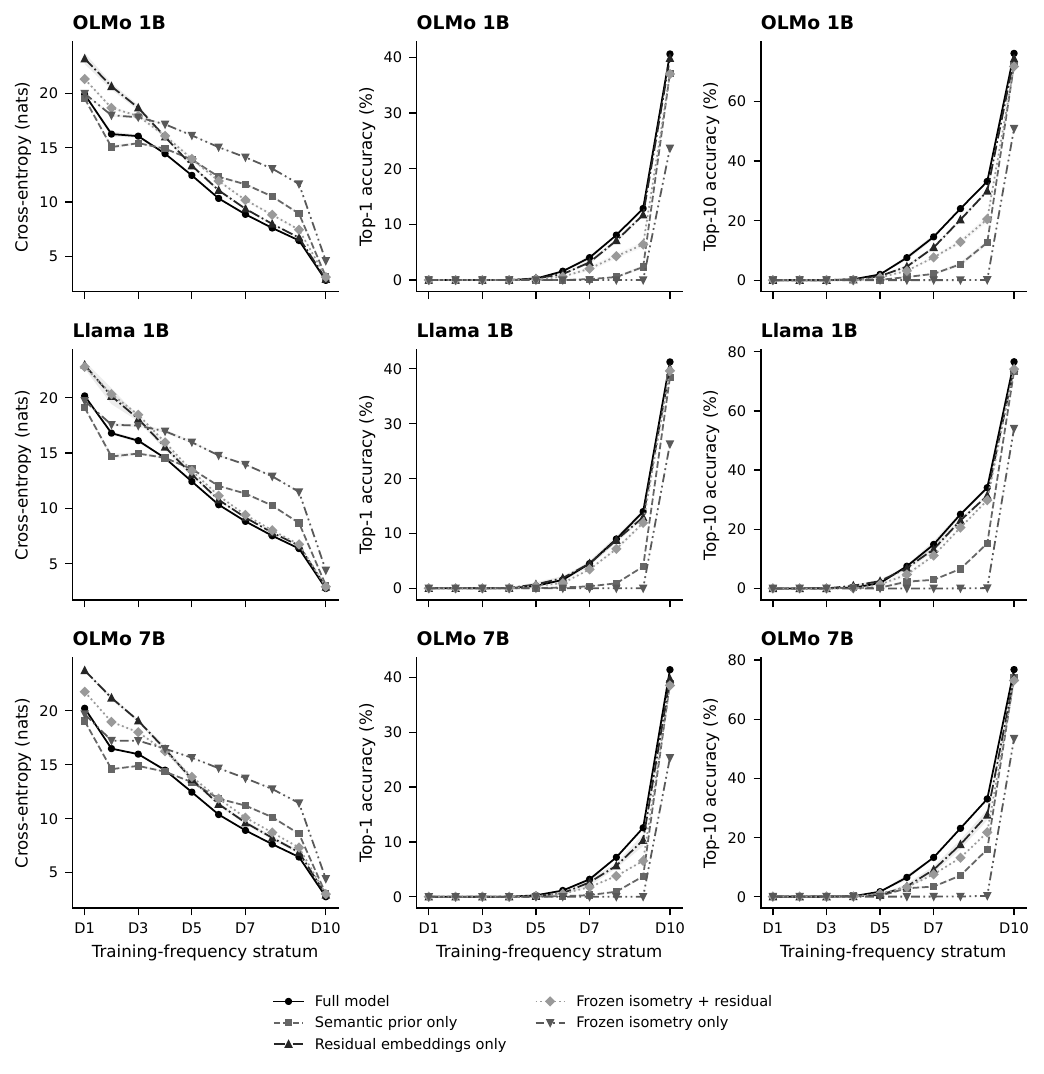}
\caption{\textbf{Absolute CPT test performance by training frequency.}
Rows correspond to OLMo2-1B, Llama3.2-1B, and OLMo2-7B; columns show CE, top-1,
and top-10 accuracy. D1 contains the rarest event types, D10 the most frequent.
Lines and bands show the mean and one sample SD across three seeds.
All metrics use the same validation-selected checkpoint within each run.}
\label{fig:decile-absolute}
\end{figure}

All results in this section use held-out test patients and, for each run, the
checkpoint selected by global validation CE; values are means and sample SDs
over three seeds. Table~\ref{tab:cpt-all-metrics} gives global CE and top-1 and
top-10 accuracy, and Figure~\ref{fig:decile-absolute} shows each metric by
frequency stratum for every configuration and backbone.

Frequency strata are deciles of event types ranked by training-split frequency
(Appendix~\ref{app:data}). Each stratum's CE is averaged over its test targets.
Ablation effects are the ablation's CE minus the Full model's CE within the
same seed. SDs reflect variability across training seeds, not uncertainty from
sampling test patients.

\begin{table}[htbp]
\centering\small
\setlength{\tabcolsep}{3pt}
\caption{Global held-out CPT test metrics for \textbf{Full model} and ablations. Mean $\pm$ sample SD over three seeds per configuration}
\label{tab:cpt-all-metrics}
\begin{tabular}{@{}lrrr@{}}
\toprule
Configuration & CE & Top-1 (\%) & Top-10 (\%)\\
\midrule
\multicolumn{4}{l}{\textit{OLMo2-1B}}\\
Full model & $2.847\pm0.006$ & $40.201\pm0.151$ & $75.433\pm0.077$\\
Semantic prior only & $3.217\pm0.019$ & $36.653\pm0.269$ & $71.378\pm0.246$\\
Residual embeddings only & $2.959\pm0.012$ & $39.431\pm0.126$ & $73.649\pm0.154$\\
Frozen isometry + residual & $3.158\pm0.027$ & $36.568\pm0.412$ & $70.913\pm0.419$\\
Frozen isometry only & $4.689\pm0.005$ & $23.342\pm0.083$ & $50.138\pm0.108$\\
\midrule
\multicolumn{4}{l}{\textit{Llama3.2-1B}}\\
Full model & $2.806\pm0.011$ & $40.857\pm0.118$ & $75.984\pm0.168$\\
Semantic prior only & $3.129\pm0.006$ & $37.948\pm0.164$ & $72.589\pm0.068$\\
Residual embeddings only & $2.937\pm0.014$ & $39.713\pm0.208$ & $73.955\pm0.145$\\
Frozen isometry + residual & $2.974\pm0.015$ & $39.236\pm0.220$ & $73.477\pm0.206$\\
Frozen isometry only & $4.500\pm0.017$ & $25.939\pm0.176$ & $53.331\pm0.257$\\
\midrule
\multicolumn{4}{l}{\textit{OLMo2-7B}}\\
Full model & $2.799\pm0.003$ & $40.941\pm0.049$ & $76.108\pm0.037$\\
Semantic prior only & $3.090\pm0.008$ & $38.401\pm0.073$ & $73.151\pm0.129$\\
Residual embeddings only & $2.976\pm0.050$ & $39.267\pm0.675$ & $73.375\pm0.771$\\
Frozen isometry + residual & $3.057\pm0.001$ & $38.054\pm0.009$ & $72.342\pm0.038$\\
Frozen isometry only & $4.513\pm0.014$ & $24.967\pm0.200$ & $52.699\pm0.246$\\
\bottomrule
\end{tabular}
\end{table}

\section{Permutation control}
\label{app:permutation}

\begin{figure}[htbp]
\centering
\includegraphics[width=\linewidth]{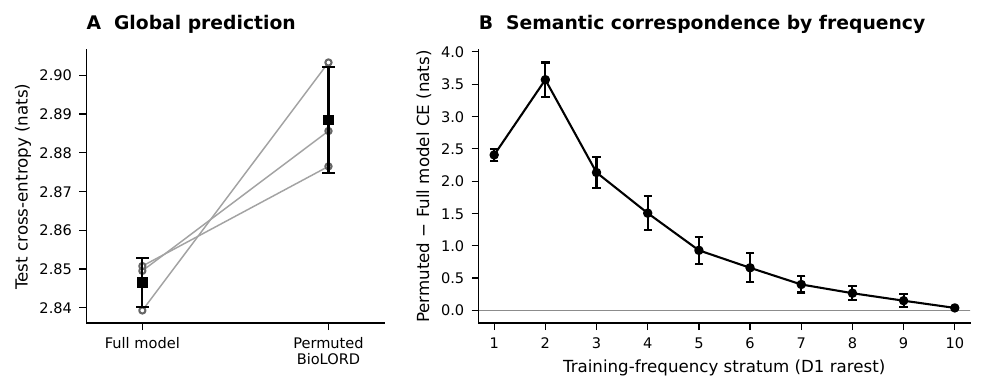}
\caption{\textbf{Meaningful BioLORD assignments improve OLMo2-1B test CE.}
(A) Global CE: gray lines connect matched training seeds; black squares
and error bars show condition means and sample SDs.
(B) Permuted minus Full-model CE by training-frequency stratum. Points and
error bars show means and sample SDs of the three paired differences.
Positive values favor Full model. D1 is rarest and D10 most frequent.}
\label{fig:permutation-ce}
\end{figure}

On OLMo2-1B, we compare the \textit{Full model} with the same model after shuffling
BioLORD vectors among events of the same modality. The control keeps the set
of vectors in each modality and the architecture unchanged, but breaks the
correspondence between each event and its clinical description. Unlike
removing the semantic pathway, it keeps the projection parameters. Results use
held-out test CE at validation-selected checkpoints (Figure~\ref{fig:permutation-ce}).

Global test CE is $2.847\pm0.006$ for the Full model and $2.888\pm0.014$ for
the permuted model, a paired reduction of $0.042\pm0.020$ nats. The Full model
is better in every stratum and every seed. Its advantage is much larger in D1
and D2 ($2.406\pm0.098$ and $3.569\pm0.266$ nats) than in D10
($0.039\pm0.019$), with the largest gap in D2. The most frequent stratum holds
approximately 98.60\% of test targets, which is why these large effects on
rare events carry little weight in the global average.

This comparison supports a role for correct semantic correspondence beyond the
distribution of the vectors. It is limited to OLMo2-1B at rank 64. Selected
checkpoints differ between conditions (\textit{Full model}: 130K, 115K, and 150K steps;
permuted: 135K, 160K, and 180K), so the effect reflects the shared selection
rule rather than equal training exposure. Three seeds do not measure
uncertainty over possible permutations.

\section{Residual rank}
\label{app:rank-sweep}

\begin{figure}[htbp]
\centering
\includegraphics[width=\linewidth]{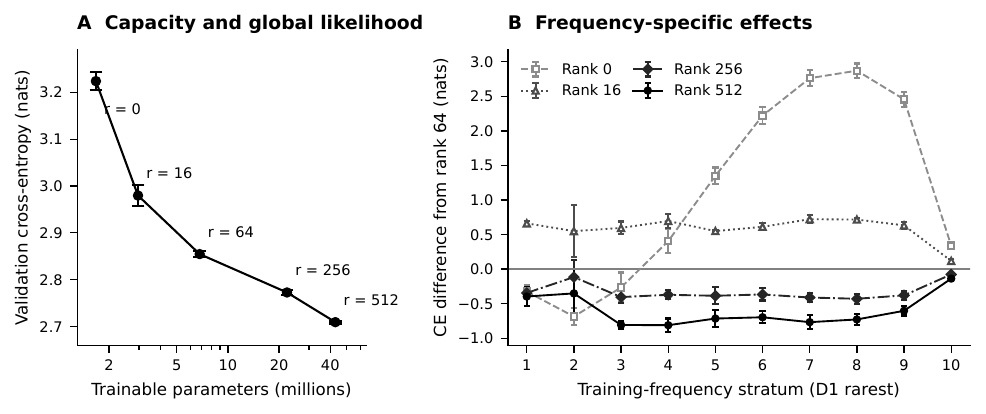}
\caption{\textbf{OLMo2-1B residual capacity and validation likelihood.}
(A) Global validation CE versus total trainable adaptation parameters on a
logarithmic parameter axis. Error bars are sample SD over three seeds.
(B) Per-stratum CE minus rank-64 CE, paired by seed; error bars show SDs of
paired differences. Negative values improve on rank 64. D1 is rarest and D10 most frequent.}
\label{fig:rank-tradeoff}
\end{figure}

On OLMo2-1B, we keep the semantic pathway and compare residual ranks 0, 16,
64, 256, and 512, with three seeds each (Figure~\ref{fig:rank-tradeoff}). This
sweep characterizes a capacity trade-off on validation data and is reported
separately from the held-out test results.

Global validation CE decreases from 3.224 at rank 0 to 2.709 at rank 512, with
the same ordering in each seed. Rank 64 uses 6.82M trainable parameters versus
42.9M at rank 512, about 16\% as many, while capturing 72\% of the global
improvement from rank 0 to rank 512. Increasing rank therefore improves
aggregate fit, with diminishing gains per added parameter. The stratified
curves qualify this pattern: larger ranks improve many middle and frequent
strata, but the two rarest strata do not improve consistently with rank, and
rank 0 remains competitive in the extreme tail despite its much worse global
CE. Extra residual capacity therefore does not guarantee better prediction
across the whole vocabulary.

\section{Downstream results}
\label{app:downstream-results}

This section gives the full downstream results summarized in
Table~\ref{tab:downstream-components}: paired component effects with SDs
(Table~\ref{tab:downstream-paired-sd}) and absolute scores for every
configuration (Tables~\ref{tab:downstream-absolute-infectious}
and~\ref{tab:downstream-absolute-syndromic}).

\paragraph{Class views.}
All 14, Diseases 13, Frequent 7, and Rare 6 average one-versus-rest class
metrics from the same 14-way model. The views change which classes enter the
macro average, not which patients are evaluated; in particular, Rare 6 is
neither a separately trained classifier nor an evaluation restricted to
rare-disease patients. Diseases 13 equals
$(7\,\mathrm{AUPRC}_{7} + 6\,\mathrm{AUPRC}_{6})/13$. The GI and respiratory
scores are one-versus-rest scores from the four-way syndromic model, and
GI/Resp. in Table~\ref{tab:downstream-components} averages the two.

\paragraph{Averaging and uncertainty.}
For each seed, metrics are averaged with equal weight over 2020, 2021, and
2022, and we report the mean and sample SD of these seed averages. Paired
effects subtract ablation scores from \textit{Full model} scores within each seed and
year before the same aggregation. SDs describe variability across training
seeds, not confidence intervals for new patients, and no significance tests
are computed. Baselines do not depend on CPT seeds and are run once, so they
have no SD.

\paragraph{Interpreting component effects.}
Comparing the \textit{Full model} with \emph{Residual embeddings only} removes shared
semantic information together with the projection parameters; comparing with
\emph{Frozen isometry + residual} keeps the descriptions and residual but
removes projection learning; and comparing with \emph{Semantic prior only}
removes the event-specific residual. The permutation control was run for CPT
only (Appendix~\ref{app:permutation}), so the downstream results do not
isolate the effect of correct event-to-description correspondence. Component
effects also need not add up. For Rare 6, the residual's benefit under a
learned projection minus its benefit under a fixed projection is $-3.27$,
$-2.82$, and $-3.88$ AUPRC points for OLMo2-1B, Llama3.2-1B, and OLMo2-7B:
the residual helps less when the projection is learned, so the two components
partly compensate for each other.


\begin{table}[htbp]
\centering\small
\setlength{\tabcolsep}{3pt}
\caption{Paired downstream advantages of \textit{Full model} over ablations in AUPRC percentage points (mean $\pm$ sample SD of three seed-level differences).}
\label{tab:downstream-paired-sd}
\begin{tabular}{@{}lrrr@{}}
\toprule
Configuration & Frequent 7 & Rare 6 & GI/Resp.\\
\midrule
\multicolumn{4}{l}{\textit{OLMo2-1B}}\\
Semantic prior only & $+0.38\pm0.16$ & $+1.91\pm1.14$ & $+0.30\pm0.53$\\
Residual embeddings only & $+1.30\pm0.84$ & $+3.60\pm1.18$ & $+0.66\pm0.79$\\
Frozen isometry + residual & $+2.69\pm2.01$ & $+3.30\pm2.70$ & $+6.69\pm2.03$\\
Frozen isometry only & $+5.95\pm0.22$ & $+8.48\pm1.35$ & $+6.78\pm2.33$\\
\midrule
\multicolumn{4}{l}{\textit{Llama3.2-1B}}\\
Semantic prior only & $+0.54\pm0.39$ & $+0.86\pm1.02$ & $+1.82\pm0.71$\\
Residual embeddings only & $+2.16\pm0.75$ & $+4.16\pm0.93$ & $+1.95\pm2.05$\\
Frozen isometry + residual & $+1.43\pm0.19$ & $+3.95\pm0.45$ & $+1.66\pm1.13$\\
Frozen isometry only & $+4.43\pm0.58$ & $+7.64\pm0.34$ & $+3.15\pm0.79$\\
\midrule
\multicolumn{4}{l}{\textit{OLMo2-7B}}\\
Semantic prior only & $+1.06\pm0.49$ & $+2.08\pm0.55$ & $+0.39\pm0.62$\\
Residual embeddings only & $+2.47\pm0.63$ & $+5.10\pm1.26$ & $+0.72\pm0.80$\\
Frozen isometry + residual & $+1.47\pm0.50$ & $+3.21\pm0.62$ & $+1.61\pm1.46$\\
Frozen isometry only & $+6.52\pm0.72$ & $+9.17\pm0.54$ & $+4.33\pm0.47$\\
\bottomrule
\end{tabular}
\end{table}

\begin{table}[htbp]
\centering\small
\setlength{\tabcolsep}{3pt}
\caption{Infectious-disease classification: absolute macro-AUPRC values for the \textit{Full model} and its ablations. Values are mean $\pm$ sample SD across three seed-level annual averages.}
\label{tab:downstream-absolute-infectious}
\begin{tabular}{lrrrr}
\toprule
Configuration & All 14 & Diseases 13 & Frequent 7 & Rare 6\\
\midrule
\multicolumn{5}{l}{\textit{OLMo2-1B}}\\
Full model & $0.366\pm0.004$ & $0.326\pm0.005$ & $0.405\pm0.002$ & $0.233\pm0.010$\\
Semantic prior only & $0.355\pm0.002$ & $0.315\pm0.003$ & $0.402\pm0.001$ & $0.214\pm0.005$\\
Residual embeddings only & $0.344\pm0.005$ & $0.302\pm0.006$ & $0.392\pm0.007$ & $0.197\pm0.004$\\
Frozen isometry + residual & $0.338\pm0.020$ & $0.296\pm0.021$ & $0.379\pm0.019$ & $0.200\pm0.024$\\
Frozen isometry only & $0.298\pm0.001$ & $0.255\pm0.001$ & $0.346\pm0.001$ & $0.148\pm0.004$\\
\midrule
\multicolumn{5}{l}{\textit{Llama3.2-1B}}\\
Full model & $0.380\pm0.002$ & $0.341\pm0.002$ & $0.421\pm0.002$ & $0.247\pm0.004$\\
Semantic prior only & $0.373\pm0.004$ & $0.334\pm0.005$ & $0.416\pm0.002$ & $0.239\pm0.012$\\
Residual embeddings only & $0.351\pm0.010$ & $0.310\pm0.010$ & $0.400\pm0.008$ & $0.206\pm0.013$\\
Frozen isometry + residual & $0.355\pm0.001$ & $0.315\pm0.001$ & $0.407\pm0.003$ & $0.208\pm0.005$\\
Frozen isometry only & $0.324\pm0.003$ & $0.282\pm0.003$ & $0.377\pm0.006$ & $0.171\pm0.000$\\
\midrule
\multicolumn{5}{l}{\textit{OLMo2-7B}}\\
Full model & $0.378\pm0.004$ & $0.339\pm0.004$ & $0.421\pm0.006$ & $0.242\pm0.009$\\
Semantic prior only & $0.363\pm0.003$ & $0.323\pm0.003$ & $0.410\pm0.001$ & $0.222\pm0.006$\\
Residual embeddings only & $0.343\pm0.001$ & $0.302\pm0.001$ & $0.396\pm0.001$ & $0.191\pm0.003$\\
Frozen isometry + residual & $0.356\pm0.003$ & $0.316\pm0.004$ & $0.406\pm0.002$ & $0.210\pm0.005$\\
Frozen isometry only & $0.304\pm0.003$ & $0.261\pm0.002$ & $0.356\pm0.003$ & $0.151\pm0.006$\\
\bottomrule
\end{tabular}
\end{table}

\begin{table}[htbp]
\centering\small
\setlength{\tabcolsep}{3pt}
\caption{Syndromic classification: absolute macro-AUPRC (multiclass) and AUPRC (GI and Respiratory) values for the \textit{Full model} and its ablations. Values are mean $\pm$ sample SD across three seed-level annual averages.}
\label{tab:downstream-absolute-syndromic}
\begin{tabular}{lrrr}
\toprule
Configuration & All 4 & GI & Respiratory\\
\midrule
\multicolumn{4}{l}{\textit{OLMo2-1B}}\\
Full model & $0.791\pm0.005$ & $0.681\pm0.005$ & $0.730\pm0.009$\\
Semantic prior only & $0.787\pm0.003$ & $0.679\pm0.008$ & $0.726\pm0.004$\\
Residual embeddings only & $0.786\pm0.000$ & $0.670\pm0.003$ & $0.727\pm0.001$\\
Frozen isometry + residual & $0.742\pm0.022$ & $0.649\pm0.008$ & $0.628\pm0.046$\\
Frozen isometry only & $0.742\pm0.013$ & $0.596\pm0.020$ & $0.679\pm0.020$\\
\midrule
\multicolumn{4}{l}{\textit{Llama3.2-1B}}\\
Full model & $0.797\pm0.003$ & $0.686\pm0.005$ & $0.741\pm0.004$\\
Semantic prior only & $0.783\pm0.003$ & $0.677\pm0.006$ & $0.713\pm0.006$\\
Residual embeddings only & $0.783\pm0.017$ & $0.665\pm0.018$ & $0.723\pm0.031$\\
Frozen isometry + residual & $0.783\pm0.010$ & $0.670\pm0.006$ & $0.723\pm0.021$\\
Frozen isometry only & $0.771\pm0.009$ & $0.630\pm0.010$ & $0.733\pm0.013$\\
\midrule
\multicolumn{4}{l}{\textit{OLMo2-7B}}\\
Full model & $0.798\pm0.003$ & $0.688\pm0.005$ & $0.740\pm0.008$\\
Semantic prior only & $0.794\pm0.006$ & $0.683\pm0.015$ & $0.738\pm0.007$\\
Residual embeddings only & $0.793\pm0.001$ & $0.677\pm0.001$ & $0.737\pm0.005$\\
Frozen isometry + residual & $0.786\pm0.007$ & $0.677\pm0.008$ & $0.720\pm0.012$\\
Frozen isometry only & $0.765\pm0.005$ & $0.627\pm0.008$ & $0.715\pm0.005$\\
\bottomrule
\end{tabular}
\end{table}

\section{Training of NEP and EHRAdapt}
\label{app:nep-ehradapt-training}

  Table~\ref{tab:nep-ehradapt-training} compares the training of the NEP baselines with that of the EHRAdapt \textit{Full model} on each backbone. Both methods keep the backbone frozen
  and train only on the patient-level CPT training split, which contains 3,945,557 patients, 4.46 million chunks, and 721 million prediction targets; neither uses validation or test
  patients for training. NEP trains rank-16 LoRA adapters for 10,000 steps at a global batch of 512, yielding 5.12 million instruction-response examples drawn from 3,945,492--3,945,549
  distinct patients. The settings disclosed by \citet{chen2025buildingehrfoundationmodel}, namely the LoRA rank, optimizer, learning rate and schedule, global batch size, and number of
  steps, match the original, whereas the prompt and serialization template, which are unpublished, follow our own format. EHRAdapt trains for at most six epochs with validation-based early stopping, and every
  run touches all training patients. The three OLMo2-7B runs stopped after 100,000--145,000 optimizer steps, corresponding to 2.07--3.00 billion supervised targets and 2.87--4.16
  target-equivalent epochs, compared with 3.73--4.73 epochs for OLMo2-1B and 2.15--2.72 for Llama3.2-1B; early stopping therefore ended every run before the six-epoch maximum. Each NEP example serializes on average 58.6 context events into 1,557 BPE tokens and supervises a single target. Consequently, NEP processes 2.3--5.1 times as
  many backbone input tokens as EHRAdapt, whereas EHRAdapt receives 300--670 times as many supervised targets. GPU-hours are optimizer-loop hours, computed as elapsed time multiplied by
  the number of devices, and exclude queueing, final validation, and checkpoint export. NEP and EHRAdapt both ran on AMD Instinct MI300A GPUs (gfx942), and NEP used more devices: 32
  versus 16 for the 1B backbones and 128 versus 64 for OLMo2-7B.

  \begin{table}[htb]
    \centering\small
  \setlength{\tabcolsep}{3pt}
  \caption{Training of NEP and the EHRAdapt \textit{Full model}. Ranges span three seeds. Targets are supervised next-event predictions; input tokens are backbone input positions
  processed during training.}
  \label{tab:nep-ehradapt-training}
  \begin{tabular}{@{}llrrrrr@{}}
  \toprule
  Backbone & Method & Trainable params & Targets & Input tokens & Tokens/event & GPU-hours\\
  \midrule
  OLMo2-1B & NEP & 6.82M & 5.12M & 7.96--7.97B & $\approx$27 & 301--303\\
   & EHRAdapt & 6.82M & 2.69--3.41B & 2.69--3.41B & 1 & 144--184\\
  \midrule
  Llama3.2-1B & NEP & 6.03M & 5.12M & 7.96--7.97B & $\approx$27 & 276\\
   & EHRAdapt & 6.83M & 1.55--1.96B & 1.55--1.96B & 1 & 119--146\\
  \midrule
  OLMo2-7B & NEP & 8.39M & 5.12M & 7.96--7.98B & $\approx$27 & 1,084--1,564\\
   & EHRAdapt & 8.54M & 2.07--3.00B & 2.07--3.00B & 1 & 541--785\\
  \bottomrule
  \end{tabular}
  \end{table}

\end{document}